\documentclass{article}
\usepackage{ijcai21}

\usepackage{times}
\usepackage{soul}
\usepackage{url}
\usepackage[hidelinks]{hyperref}
\usepackage[utf8]{inputenc}
\usepackage[small]{caption}
\usepackage{graphicx}
\usepackage{amsmath}
\usepackage{amsthm}
\usepackage{booktabs}
\usepackage{algorithm}
\usepackage{algorithmic}
\usepackage{amssymb}
\usepackage{pifont}
\newcommand{\cmark}{\ding{51}}%
\newcommand{\xmark}{\ding{55}}%

\title{IMENet: Joint 3D Semantic Scene Completion and 2D Semantic Segmentation through Iterative Mutual Enhancement}

\author{
Jie Li$^{1,2}$
\and
Laiyan Ding$^1$\and
Rui Huang$^{1,2}$ \thanks{Corresponding author.}
\affiliations
$^1$The Chinese University of Hong Kong, Shenzhen\\
$^2$Shenzhen Institute of Artificial Intelligence and Robotics for Society\\
\emails
\{jieli1, laiyanding\}@link.cuhk.edu.cn,
ruihuang@cuhk.edu.cn 
}

\begin{document}
\maketitle
\begin{abstract}
  \hyphenpenalty=10000
  \allowbreak
  3D semantic scene completion and 2D semantic segmentation are two tightly correlated tasks that are both essential for indoor scene understanding, because they predict the same semantic classes, using positively correlated high-level features. Current methods use 2D features extracted from early-fused RGB-D images for 2D segmentation to improve 3D scene completion. We argue that this sequential scheme does not ensure these two tasks fully benefit each other, and present an Iterative Mutual Enhancement Network (IMENet) to solve them jointly, which interactively refines the two tasks at the late prediction stage. Specifically, two refinement modules are developed under a unified framework for the two tasks. The first is a 2D Deformable Context Pyramid (DCP) module, which receives the projection from the current 3D predictions to refine the 2D predictions. In turn, a 3D Deformable Depth Attention (DDA) module is proposed to leverage the reprojected results from 2D predictions to update the coarse 3D predictions. This iterative fusion happens to the stable high-level features of both tasks at a late stage. Extensive experiments on NYU and NYUCAD datasets verify the effectiveness of the proposed iterative late fusion scheme, and our approach outperforms the state of the art on both 3D semantic scene completion and 2D semantic segmentation.

\end{abstract}

\section{Introduction}
3D Semantic Scene Completion (SSC) has recently drawn increased attention as it aims at simultaneously producing a complete 3D voxel representation of volumetric occupancy and semantic labels from a depth image.
Driven by deep Convolutional Neural Networks (CNNs), studies on SSC have achieved significant progress and start to benefit some real-world computer vision and robotics applications, such as indoor navigation, robot grasping, and virtual reality.
However, due to the sparsity of 3D data (i.e., most voxels in a 3D scene are empty), it is a huge challenge for SSC to obtain more accurate results. 

On the other hand, 2D Semantic Segmentation (SS) can predict dense class labels from a high-resolution 2D image.
Although 2D SS is one dimension lower than 3D SSC, they share some important characteristics that enable them to benefit each other. For example, they produce the same element-wise semantic labels, and both can learn and utilize high-level semantic features under the deep learning framework. Besides, various inexpensive RGB-D imaging devices make the simultaneous acquisition of both data sources very easy. It seems desirable to consider them together so that both get improved. Intuitively, the dense prediction of 2D SS can make up the sparsity of 3D SSC, and the complete shape/layout of the 3D scene can help distinguish inseparable 2D regions.
\begin{figure}[t] 
	\center
	\includegraphics[width=\columnwidth, height=4.0cm]{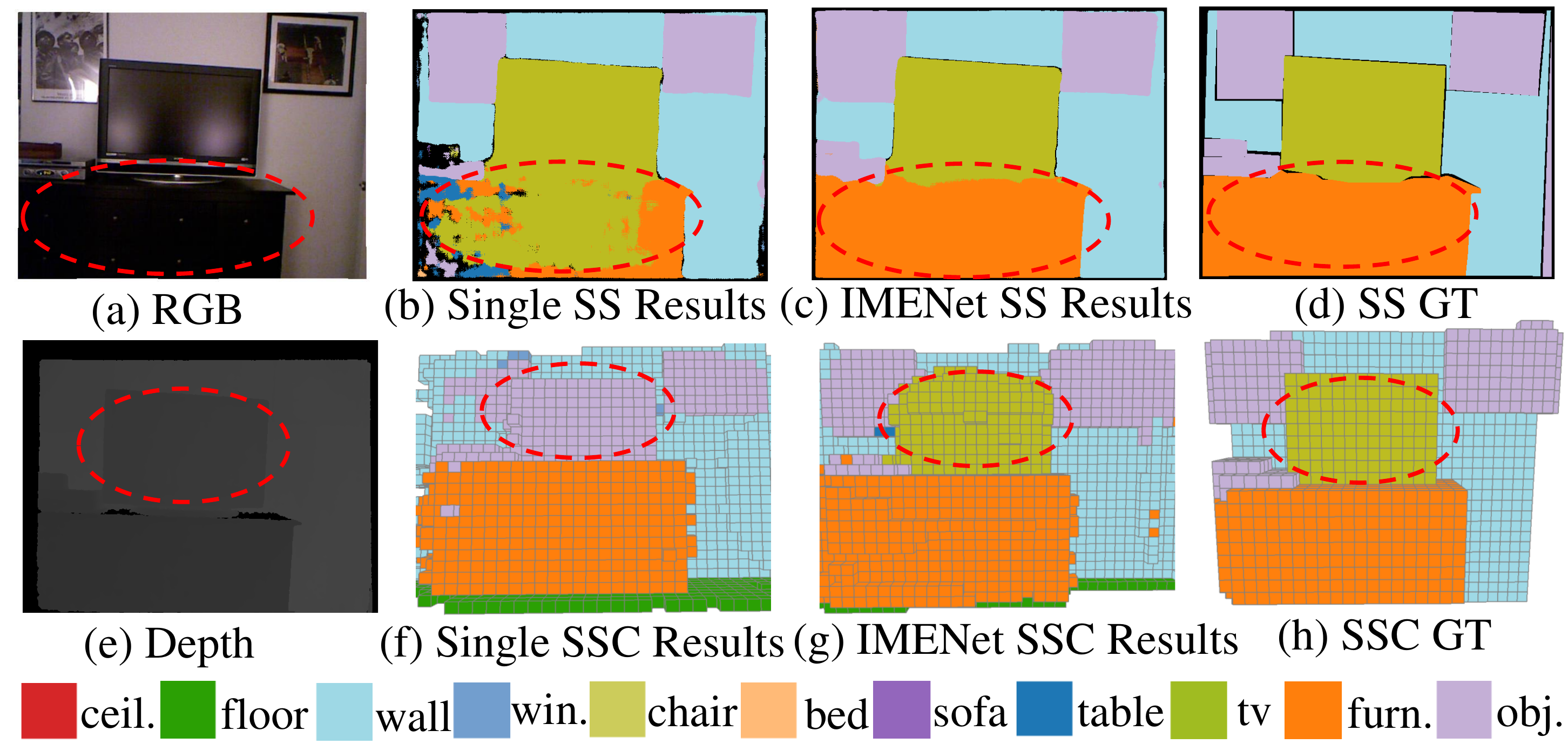}
	\caption[stereo]{\textbf{An example of mutual promotion between 3D semantic scene completion (SSC) and 2D semantic segmentation (SS) tasks}.
The proposed IMENet can improve the accuracy of each other by using the iterative complementary information of different tasks.
}
	\label{fig-intro}
\end{figure}

Several previous studies \cite{liu2018see,garbade2019two,li2020attention} introduced 2D SS into 3D SSC via joint multi-task learning, and improved the SSC accuracy.
However, these methods generally perform 2D SS before 3D SSC, and the sequential scheme not only might bring 2D SS errors into 3D SSC, but also provides no runtime feedback from 3D SSC to 2D SS.
Besides, the early/shallow fusion of RGB-D multi-modality data in the previous works cannot fully exploit the high-level features of each specific input.

To address the above problems, we propose a novel Iterative Mutual Enhancement Network (IMENet) to realize mutual enhancement between 3D SSC and 2D SS, in an iterative and bi-directional way, on the stable high-level features from the late stage of the deep neural network.
Our motivation is that geometric information from 3D SSC and high-resolution color/texture features from 2D SS are complementary, and can help the learning of each other. In addition, the iterative interaction between the two tasks allows error-correction based on gradually improved predictions in each task.

Taking the scenario shown in Figure \ref{fig-intro} as an example, because of the similar colors/textures of the \emph{furniture} and \emph{tv} categories, it is very hard for a 2D SS network to accurately separate the two classes (Figure \ref{fig-intro} (b)).
On the other hand, since the \emph{tv} and \emph{pictures} categories often have similar depth, the existing 3D SSC network may confuse these two classes (Figure \ref{fig-intro} (f)).
Both failures can be significantly avoided if we leverage the complementary information of the two tasks jointly. Specifically, our proposed IMENet can introduce the distinct geometric shapes of \emph{tv} and \emph{furniture} from the SSC branch to SS, and in turn, it can also provide different appearance semantics of \emph{tv} and \emph{pictures} from the SS branch to SSC, thus prevent both failures (Figure \ref{fig-intro} (c) and (g)).

Additionally, we believe that, even though 2D SS and 3D SSC focus on different channels of the RGB-D input and produce outputs of different dimensions, they share the same semantic labels in the end, and therefore the features learned at the late stages in both tasks are highly correlated. Instead of fusing RGB and D channels early, we adopt a late fusion strategy to fully utilize the correlation of the high-level features from both RGB and D channels.
To this end, we designed two task-shared modules named 2D Deformable Context Pyramid (DCP) and 3D Deformable Depth Attention (DDA) for implicit mutual knowledge fusion.
In particular, the pixel-wise semantic segmentation and the voxel-wise semantic completion are leveraged to maintain the global-local geometric contexts and accommodate shape and scale variations of different objects through the DCP module.
Meanwhile, to probe the inherent property of depth thoroughly, DDA adopts an attention strategy to weigh the depth relationships of different positions after applying the similar DCP structure in 3D.
Finally, through iterative interactions between DCP and DDA, our IMENet can explore more refined predictions of both SSC and SS.

In summary, our main contributions are three-fold:
1) We propose a cross-dimensional multi-task iterative mutual enhancement network named IMENet for RGB-D indoor scene understanding.
  Coarse predictions of task-specific sub-networks can be further refined by properly propagating the complementary information.
2) A late fusion strategy is designed to capture high-level specific features of multi-modality input via the 2D Deformable Context Pyramid and 3D Deformable Depth Attention modules.
3) Experimental results on NYU and NYUCAD datasets demonstrate that our IMENet outperforms the state-of-the-art methods. 
\section{Related Work}
\subsection{Semantic Segmentation}
The goal of semantic segmentation is to predict class labels for the input raw data, which can be 2D images and 3D volumes.
Although 2D dense SS has made great progress due to the rapid development of 2D CNNs \cite{chen2017deeplab}, 3D SS is easier for intelligent agents to understand because of the exact geometries \cite{qi2017pointnet,qi2017pointnet++}.
However, it is still challenging to obtain accurate and consistent semantic labels of 3D data because of the inherent sparsity.
Therefore, our work will explore interactions between 2D pixel-wise and 3D voxel-wise semantic segmentation to exploit complimentary information for mutual improvements.

\subsection{Semantic Scene Completion}
Semantic Scene Completion methods are divided into two main categories: 
(1) \textbf{Single depth-based} SSC \cite{wang2018adversarial,wang2019forknet,Zhang_2019_ICCV,li2019depth} usually reconstructs the semantic complete scenes by solely using the depth as input.
In particular, SSCNet \cite{song2017semantic} is a seminal work that uses an end-to-end dilation-based convolution network to achieve SSC. 
Although depth-based methods have made considerable progress, the absence of color and texture details always constrains their performance.
 (2) 
 \textbf{RGB-D-based} SSC can fully use the complementary information from cross-modality inputs to acquire richer features \cite{chen20203d}.
 \cite{li2019rgbd} first presented a light-weight DDRNet by decomposing the kernel on each dimension.
 Several works were later proposed to achieve dynamic fusion modes and adaptive kernel size selections \cite{liu20203d,li2020anisotropic}.

Closer to our work, \cite{liu2018see} designed a two-stage framework with the SS cascaded by a semantic completion module.
Similarly, \cite{li2020attention} directly conducted 3D semantic completion with 2D SS features as input.
In addition, \cite{garbade2019two} constructed an incomplete 3D semantic tensor from the 2D SS module 
to formulate the SSC input.  
However, the sequential arrangement of SS and SSC, as well as the early shallow fusion approaches, have their own limits as discussed in the introduction. Therefore, in this paper we will adopt an iterative and interactive structure between SS and SSC.
\subsection{Multi-task Joint Learning}
Joint learning methods with CNNs have been applied to solve multiple vision tasks simultaneously \cite{yan2020sparse}.
For instance, \cite{zhang2019joint} designed a task-recursive learning (TRL) mechanism, which shares a typical CNN encoder and performs interactions across multiple homogeneous 2D vision tasks at four different scales of the decoder.
Different from prior works, our IMENet conducts the iterative interactions between two relatively independent sub-networks, dealing with problems in different dimensions.
Besides, the weights of specific sub-networks are updated alternately, which allows every iteration to use the latest and more refined predictions from the other task.
\section{Methodology}
\begin{figure}[htp]
	\center
	\includegraphics[width=\columnwidth, height=4.25cm]{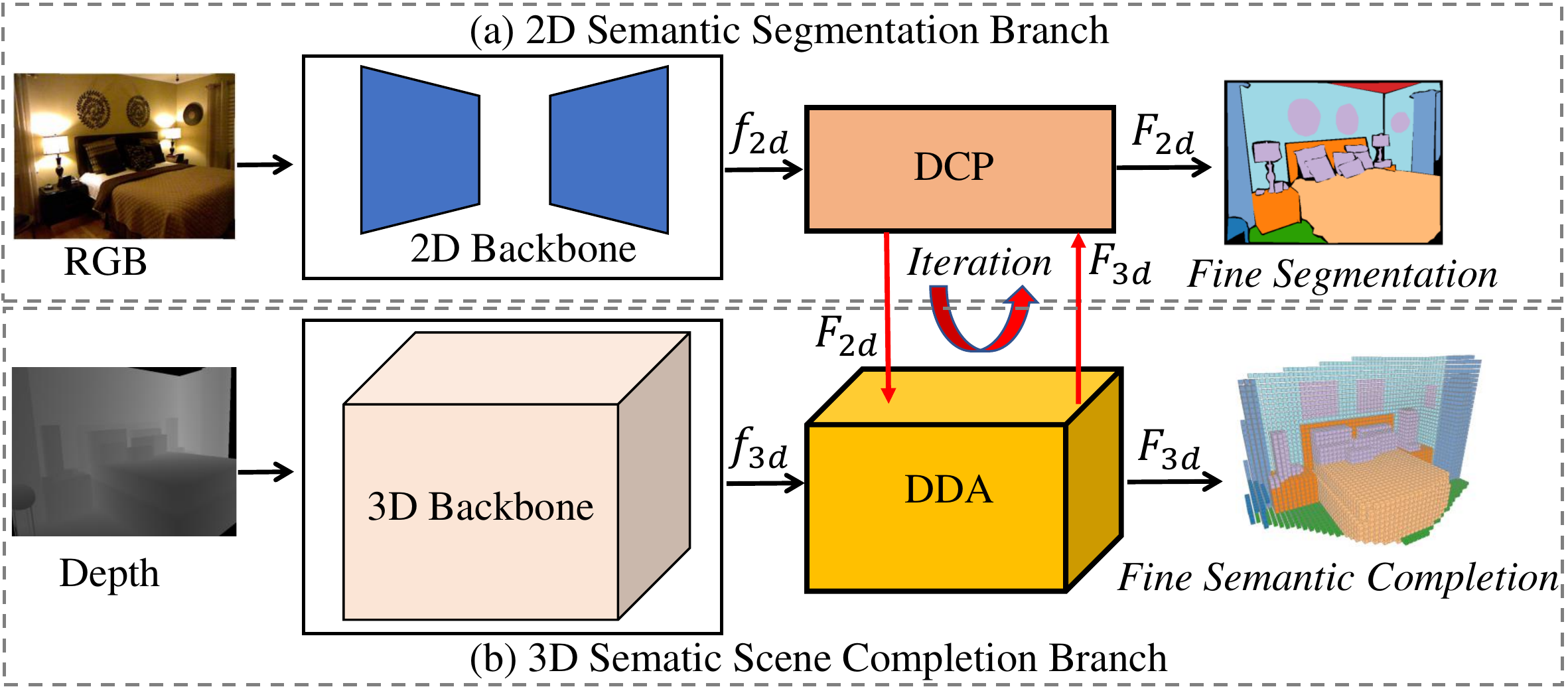}
	\caption[stereo]{The overall pipeline of our IMENet. It consists of two branches, i.e., (a) 2D semantic segmentation branch and (b) 3D sematic scene completion branch.
The red arrows indicate feature interactions between two branches.
}
	\label{fig-framework}
\end{figure}

The overall pipeline of our proposed IMENet is shown in Figure \ref{fig-framework}.
The main idea is that the 2D dense segmentation provides semantic priors for 3D semantic completion. Meanwhile, the accurate semantic
complete results contribute to producing a correct scheme for image pixel annotations accordingly.
The basic strategy is to update the 2D SS branch weights with the 3D SSC weights fixed and vice versa.
\subsection{2D Semantic Segmentation Branch}
\label{2dss}
The semantic segmentation branch aims at acquiring pixel-wise semantic predictions of a scene, as shown in Figure \ref{fig-framework} (a).
Similar to \cite{liu2018see,garbade2019two,li2020attention}, we intend to process the SS task in the 2D image space before transferring the segmentation results to the 3D volume space for time efficiency and higher output resolution.
A pre-trained encoder-decoder architecture Seg2DNet \cite{liu2018see} is applied to realize coarse SS predictions $f_{2d} \in \mathbb{R}^{12 \times W \times H}$.
The encoder is ResNet-101 \cite{he2016deep} and the decoder contains a series of dense upsampling convolutions \cite{wang2018understanding}.
After the 2D backbone, the coarse result $f_{2d}$ is sent to the Deformable Context Pyramid (DCP) module to get a further refined semantic prediction map $F_{2d} \in \mathbb{R}^{1 \times W \times H}$, where 1, \emph{W}, and \emph{H} indicate the channels, width, and height of features.
\subsection{3D Semantic Scene Completion Branch}
\label{3dssc}
The goal of the 3D semantic scene completion branch is to map every voxel in the view frustum to one of $K + 1$ labels $C = \{c_{0}, ..., c_{K}\}$ , where $K$ ($= 11$ in this paper) is the number of object classes, and $c_{0} $ represents empty voxels.
As shown in Figure \ref{fig-framework} (b), the depth input is initially fed into the 3D backbone to generate a 12-channel semantic completion feature map $f_{3d} \in \mathbb{R}^{12 \times W \times H \times D}$.
$f_{3d}$ and $F_{2d}$ are then sent into the Deformable Depth Attention (DDA) module for a better cross-modality feature aggregation.
Finally, we can obtain the refined 3D prediction $F_{3d} \in \mathbb{R}^{1 \times W \times H \times D}$, where \emph{D} indicates the depth of the features.
In addition, The 3D backbone can be replaced by other appealing CNNs.
\subsection{Iterative Mutual Enhancement Framework}
\begin{figure*}[htp]
	\center
	\includegraphics[width=1.6\columnwidth, height=1.8cm]{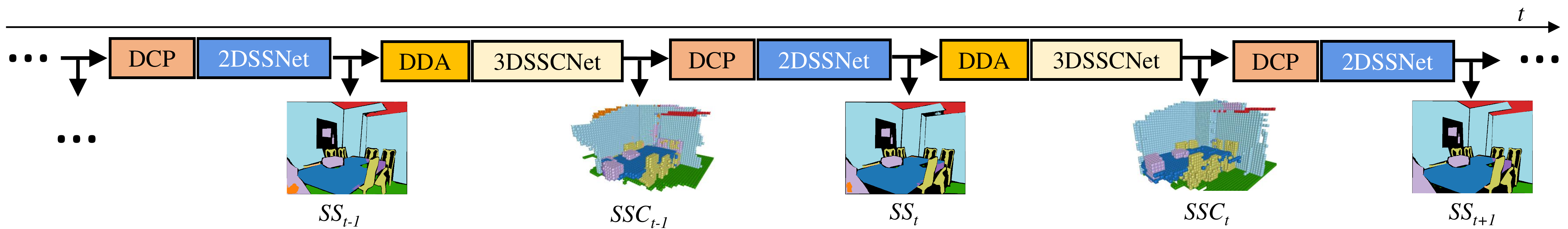}  
	\caption[stereo]{Illustration of our iterative learning. At time slice \emph{t}, the two tasks (i.e., 3D SSC and 2D SS) are progressively refined to form a task state sequence
($ SSC_{t}, SS_{t}$).
3DSSCNet and 2DSSNet are the 3D SSC branch and 2D SS branch respectively.
}
	\label{fig-iter}
\end{figure*}

To fully delve into more interactions between two tasks, this paper develops a novel iterative learning framework for closed-loop SSC and SS tasks on indoor scenes (red arrows in Figure \ref{fig-framework}).
The iteration is grounded on the assumption that the better 2D SS results should help achieve better 3D SSC performances and vice versa.
The iteration is serialized as a time axis, as shown in Figure \ref{fig-iter}. Along the time dimension $ t$, the two tasks $\{ SSC_{t}, SS_{t}\} $ mutually collaborate to boost the performance of each other.
As shown in Figure \ref{fig-framework}, given an RGB-D image as input, our IMENet first feeds the RGB and D channels into the 2D and 3D backbones respectively, to generate coarse segmentation results $f_{2d}$ and $f_{3d}$.
$f_{2d}$ and the 3D feature map $F_{3d}$ will be fed into DCP for 2D SS refinements. 
Similarly, $f_{3d}$ and $F_{2d}$ are sent to DDA for 3D task-shared fusions.
After continuous iterative learning, both the SSC and SS branches gradually improve, and finally, IMENet will output the optimal results $F_{2d}$ and $F_{3d}$ respectively.
\subsection{Deformable Context Pyramid}
\label{2dfusion}
To incorporate the geometric contextual information and accommodate object shape and scale variations, we introduce a Deformable Context Pyramid (DCP) module in the SS branch of IMENet.
In particular, inspired by the outstanding performances of the Atrous Spatial Pyramid Poolings (ASPP) \cite{chen2017deeplab}, we apply a context pyramid structure to aggregate multi-scale contexts and local geometry details to enhance the shape recovery capability of our proposed IMENet. Besides, to improve the geometric transformation modeling capacity, we take advantage of the deformable CNNs \cite{dai2017deformable} to keep the shape smooth and consistent.

\begin{figure}[htp]
	\center
	\includegraphics[width=0.76\columnwidth, height=3.25cm]{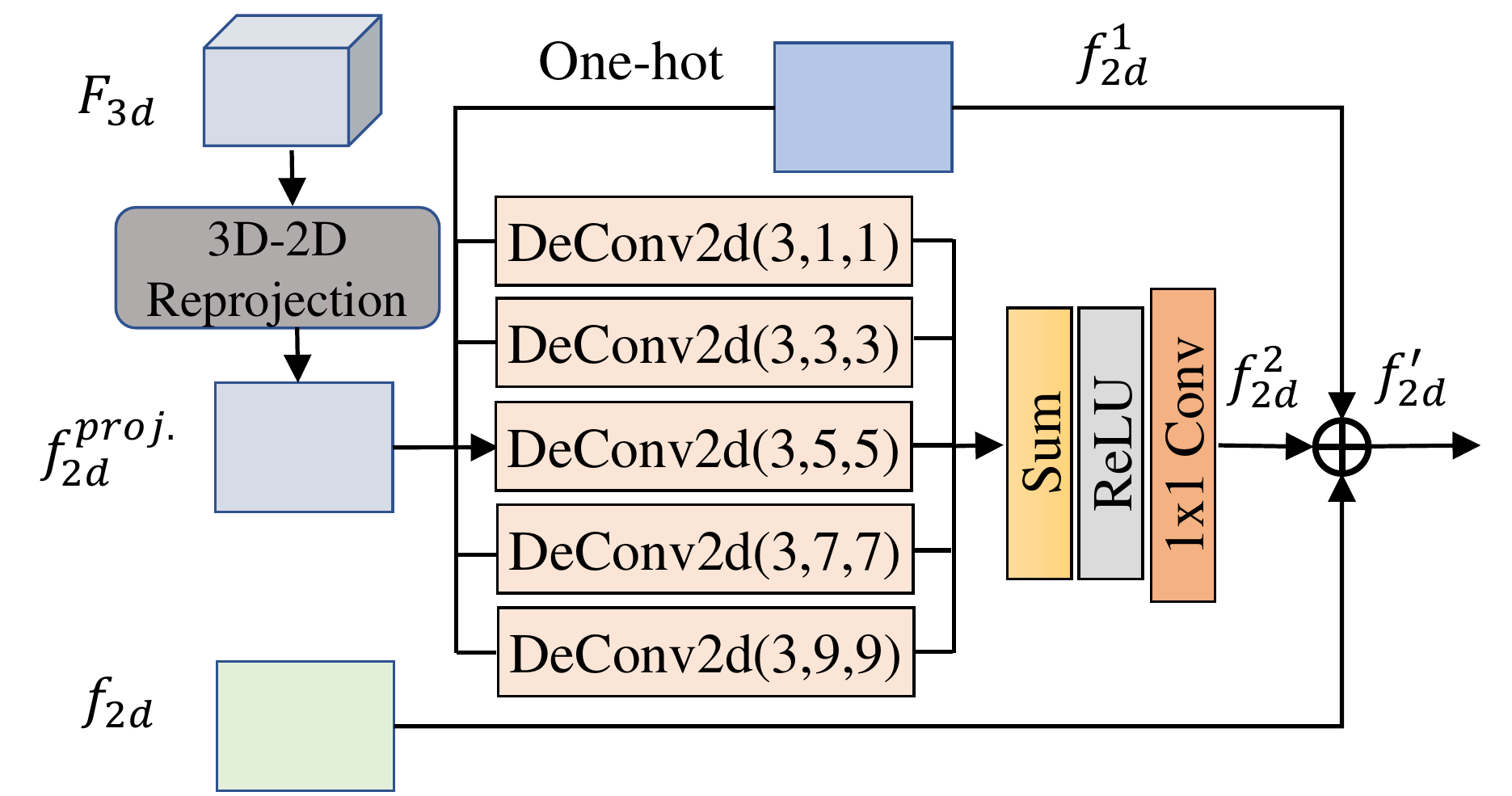}
	\caption[stereo]{The inner structure of 2D Deformable Context Pyramid.
}
	\label{fig-2dfusion}
\end{figure}

Figure \ref{fig-2dfusion} gives the details of DCP. 
Given the input $F_{3d}$ and $f_{2d}$, we first apply the 3D-2D reprojection layer to project the 3D feature volume $F_{3d}$ into a 2D feature map $f_{2d}^{proj.} \in \mathbb{R}^{1 \times W \times H}$. Then we utilize the one-hot function and deformable context pyramid to transform $f_{2d}^{proj.}$ into two new feature maps $f_{2d}^{1}, f_{2d}^{2} \in \mathbb{R}^{12 \times W \times H}$, respectively.
At last, $f_{2d}$, $f_{2d}^{1}$, and $f_{2d}^{2}$ are element-wisely added to generate a new feature map $f_{2d}^{'}$. The operation is defined as follows:
\begin{equation}\label{dcp}
\begin{split}
  f_{2d}^{'} &= DCP_{2d}(f_{2d},F_{3d})  \\
  &= f_{2d} \oplus h(p(F_{3d})) \oplus CR[\sum_{i} D_{i} ( p(F_{3d}) )],
\end{split} 
\end{equation}
where $p, h$ denote the reprojection and one-hot functions respectively, and $\oplus$ means the element-wise sum operation.
$C, D_{i}, R$ represent the 2D standard, the \emph{i}th deformable CNNs and ReLU operations.

\subsection{Deformable Depth Attention}
\label{3dfusion}
\begin{figure*}[htp]
	\center
	\includegraphics[width=1.5\columnwidth, height=3.2cm]{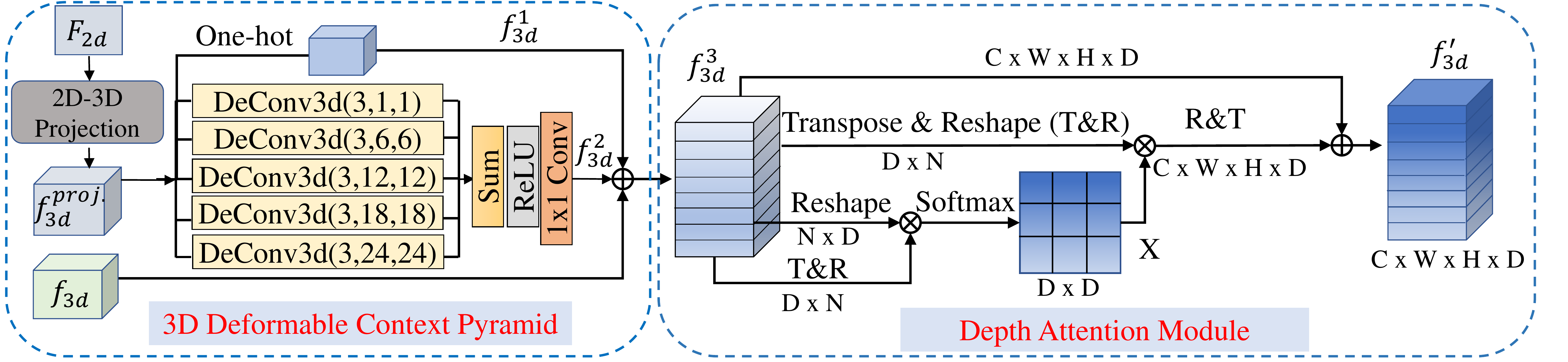}
	\caption[stereo]{The details of 3D Deformable Depth Attention Module. The kernel, padding and dilation are indicated in the convolutional layers.
}
	\label{fig-3dfusion}
\end{figure*}

The detailed structure of our Deformable Depth Attention module is illustrated in Figure \ref{fig-3dfusion}.
Due to the similarity between the SSC and SS tasks, we continue to use a similar structure $DCP_{3d}$ as the first part of our DDA module.
The main difference is that we conduct the DCP operation in 3D.
In addition, we have empirically observed that voxels belonging to the same semantic category often have a similar or consistent depth, and different objects generally have truncated depths.
Therefore, we introduce the depth attention module as the second part of our DDA module to weigh the depth importance of different categories. DDA is formulated as: 
\begin{equation}
\left\{
\begin{aligned}
& f_{3d}^{3} = DCP_{3d}(f_{3d},F_{2d}),  \\
& f_{3d,j}^{'} = \sum_{i=1}^{D}( \frac{exp(f_{3d,i}^{3}\cdot f_{3d,j}^{3})}{\sum_{i=1}^{D} exp(f_{3d,i}^{3} \cdot f_{3d,j}^{3})}) f_{3d,i}^{3} + f_{3d,j}^{3}, \\
\end{aligned}
\right.
\end{equation}
where $f_{3d}^{3}$ is the output of $DCP_{3d}$, and $f_{3d,j}^{'}$ is a feature vector in the output feature map $f_{3d}^{'}$ of DDA at the position $j$.
Different from the traditional channel attention module \cite{fu2019dual}, we directly calculate the depth attention map $X \in \mathbb{R}^{D \times D}$  from the original features $f_{3d}^{3} \in \mathbb{R}^{12 \times W \times H \times D}$.
From Equation (2), it is obvious that the final feature $f_{3d}^{'}$ is a weighted
sum of the features of all depth-channels and original features,
which models the long-range depth dependencies between feature maps.
\subsection{Loss and Training Protocol}
\textbf{Loss} To train our IMENet, the softmax cross-entropy loss is employed as in \cite{song2017semantic}. 
During training, we remove all empty voxels and do not apply the data balancing
scheme in \cite{song2017semantic} in our training process.

\noindent \textbf{Training Protocol} IMENet is implemented with PyTorch and is trained from scratch.
The cross-entropy is minimized by using the standard SGD with a momentum 0.9 and a weight decay 1e-4.
The model is trained with 4 RTX 2080Ti GPUs with batch size 4 and initial learning rate 0.005.
%
\section{Experiments}

\subsection{Dataset and Metrics}
We evaluate our method on the NYU and NYUCAD datasets. The NYU dataset is a large-scale RGB-D indoor dataset \cite{silberman2012indoor},
which contains 795/654 (training/testing) aligned pairs of RGB and depth images.
Besides, each image is also annotated with a 3D semantic label according to the ground truth completion and segmentation results from \cite{guo2015predicting}. 
However, some manually labeled voxels and their corresponding depths are not well aligned in the NYU dataset.
Therefore, we use the NYUCAD dataset \cite{firman2016structured} to tackle these misalignments, in which the depth is projected from the 3D labeled volume and thus can have higher quality.  
The voxel-level intersection over union (IoU) between the predicted and ground truth labels is used as the evaluation metric.
\subsection{Comparison with State-of-the-Arts}

Because the 3D backbone used in the 3D SSC branch is flexible and can be replaced by other appealing networks,
we conduct experiments on NYUCAD and NYU datasets with two typical networks, i.e., the pioneering SSCNet \cite{song2017semantic}, and the current state of the art TorchSSC \cite{chen20203d}, to validate the generalization ability of the proposed framework.
The results are reported in Table \ref{tab:complt_nyucad} and \ref{tab:complt_nyu}, respectively.
In Table \ref{tab:complt_nyucad}, we set SSCNet as the 3D backbone. 
It is obvious that our method significantly outperforms other methods except CCPNet \cite{Zhang_2019_ICCV} and TorchSSC in the overall accuracy (mIoU) in the SSC task.
The higher accuracy of CCPNet comes from restored details by using full resolution $240 \times 144 \times 240$, while our method just applies to a quarter resolution.
Besides, TorchSSC leverages the newly generated 3D sketch priors to guide the SSC task, 
and since the related data are only released for the NYU dataset, we only use it as our 3D backbone in the experiments on the NYU dataset.
Despite that, our IMENet can still obtain competitive results with the relatively early SSCNet as our 3D backbone.
Our IMENet achieves 7.5\% mIoU better than the baseline SSCNet, and also 1.5\% higher than the similar SS-based TS3D.
\begin{table}[htp]
		\small
		\renewcommand\tabcolsep{0.5pt}
		\begin{center}
			\resizebox{0.342\textwidth}{!}{%
				\begin{tabular}{l|ccc|c}
					\hline
					Methods&
					\ prec.&
					recall&
					IoU&	
					\ avg. \\
					\hline
					Zheng \textit{et al}. ~\cite{zheng2013beyond}\  &\ 60.1 &46.7 &34.6\  &$ -$\\
					Firman \textit{et al}.~\cite{firman2016structured} \ &\ 66.5 &69.7 &50.8\  &$ -$\\
					VVNet ~\cite{guo2018view}\  &\ 86.4 &92.0 &80.3\  &$ -$\\		
					DDRNet ~\cite{li2019rgbd}*\  &\ 88.7 &88.5 &79.4 \  &\ 42.8\\	
					GRFNet \cite{liu20203d}*\  &\  87.2 &91.0& 80.1\  &\ 45.3\\	
					AIC-Net ~\cite{li2020anisotropic}*\  &\ 88.2 &90.3 &80.5\  &\ 45.8\\	
					PALNet ~\cite{li2019depth}\  &\ 87.2 &91.7 &80.8\  &\ 46.6\\					
					CCPNet ~\cite{Zhang_2019_ICCV} &\textbf{91.3} &92.6 &82.4  &53.2\\
					TorchSSC ~\cite{chen20203d}* &90.6 &92.2 &\textbf{84.2} &\textbf{55.2}\\
                    \hline
                    SSCNet ~\cite{song2017semantic}\ &\ 75.4 &\textbf{96.3} &73.2\  &\ 40.0\\
                    TS3D ~\cite{garbade2019two}* \ &$-$ &$-$ &76.1\  &\ 46.0\\
                    Ours*  & \ \underline{84.8} &92.3 &\underline{79.1}\  & \ \underline{47.5} \\
					\hline
			\end{tabular}}
		\end{center}
        \caption{Semantic scene completion results on the NYUCAD dataset with SSCNet as the 3D backbone, and $*$ are RGB-D based methods. \textbf{Bold} numbers represent the best scores and \underline{underline} marks results that higher than similar methods.}
		\label{tab:complt_nyucad}
	\end{table}

From Table \ref{tab:complt_nyu} we can see that IMENet maintains the performance advantages and outperforms the 3D backbone TorchSSC by 3.1\% SSC mIoU and 0.8\% SC IoU.
Furthermore, IMENet-2DGT, serving as an upper bound for our IMENet, denotes the situation that SS's ground truth is used as the input to the 3D branch.
Our method tends to obtain better performance in some categories with more extreme shape variations, e.g., \emph{chair}, \emph{table}, and \emph{objects}.  
We argue that this improvement is due to our novel iterative mutual enhancements between the SS-SSC two-branch architecture, enabling SSC to use the 2D semantic prior to the full.
	\begin{table*}[htp]
		\small
		\renewcommand\tabcolsep{2.5pt}
		\begin{center}
			\resizebox{0.709\textwidth}{!}{%
				\begin{tabular}{l|ccc|ccccccccccc|c}
					\hline
					\multicolumn{1}{c|}{}
					& \multicolumn{3}{c|}{Scene Completion}
					&\multicolumn{12}{c}{Semantic Scene Completion} \\
					\hline
					Methods&
					prec.&
					recall&
					IoU&	
					ceil.&
					floor&
					wall&
					win.&
					chair&
					bed&
					sofa&
					table&
					tvs&
					furn.&
					objs.&
					avg. \\
					
					\hline
					ESSCNet ~\cite{zhang2018efficient}&71.9 &71.9 &56.2 &17.5 &75.4 &25.8 &6.7 &15.3 &53.8 &42.4 &11.2 &0.0 &33.4 &11.8 &26.7\\
					DDRNet ~\cite{li2019rgbd}* &71.5 &80.8 &61.0 &21.1 &92.2 &33.5 &6.8 &14.8 &48.3 &42.3 &13.2 &13.9 &35.3 &13.2 &30.4\\					
					SSCNet ~\cite{song2017semantic}& 59.3 &\textbf{92.9} &56.6 &15.1 &94.6 &24.7 &10.8 &17.3 &53.2 &45.9 &15.9 &13.9 &31.1 &12.6 &30.5\\
					VVNetR-120 ~\cite{guo2018view} & 69.8 &83.1 &61.1 &19.3 &94.8 &28.0 &12.2 &19.6 &57.0 &50.5 &17.6 &11.9 &35.6 &15.3 &32.9\\				
					GRFNet \cite{liu20203d}* & 68.4 &85.4 &61.2 &24.0 &91.7 &33.3 &19.0 &18.1 &51.9 &45.5 &13.4 &13.3 &37.3 &15.0 &32.9\\
					AMFNet \cite{li2020attention}* & 67.9 &82.3 &59.0 &16.7 &89.2 &27.3 &19.2 &20.2 &56.1 &50.4 &15.1 &13.5 &36.8 &18.0 &33.0\\	
					AIC-Net ~\cite{li2020anisotropic}* &62.4 &91.8 &59.2 &23.2 &90.8 &32.3 &14.8 &18.2 &51.1 &44.8 &15.2 &22.4 &38.3 &15.7 &33.3\\			
					TS3D ~\cite{garbade2019two}* &$-$ &$-$ &60.0 &9.7 &93.4 &25.5 &21.0 &17.4 &55.9 &49.2 &17.0 &27.5 &39.4 &19.3 &34.1\\	
					PALNet ~\cite{li2019depth} &68.7 &85.0 &61.3 &23.5 &92.0 &33.0 &11.6 &20.1 &53.9 &48.1 &16.2 &24.2 &37.8 &14.7 &34.1\\					
					SATNet-TNetFuse \cite{liu2018see}* & 67.3 &85.8 &60.6 &17.3 &92.1 &28.0 &16.6 &19.3 &57.5 &53.8 &17.2 &18.5 &38.4 &18.9 &34.4\\
					ForkNet ~\cite{wang2019forknet} & $-$&$-$& 63.4 &36.2 &93.8 &29.2 &18.9 &17.7 &\textbf{61.6} &52.9 &23.3 &19.5 &45.4 &20.0 &37.1\\
                    CCPNet ~\cite{Zhang_2019_ICCV} &74.2 &90.8 &63.5 &23.5 &\textbf{96.3} &35.7 &20.2 &25.8 &61.4 &\textbf{56.1} &18.1 &\textbf{28.1} &37.8 &20.1 &38.5\\
					TorchSSC ~\cite{chen20203d}* &85.0 &81.6 &71.3 &43.1 &93.6 &40.5 &24.3  &30.0 &57.1 &49.3 &29.2 &14.3 &42.5 &28.6 &41.1\\
                    \hline
                    IMENet-SS (Ours)* &\underline{\textbf{90.0}} &78.4 &\underline{\textbf{72.1}} & \underline{\textbf{43.6}} & \underline{93.6} & \underline{\textbf{42.9}} & \underline{\textbf{31.3}} & \underline{\textbf{36.6}} & \underline{57.6} & 48.4 & \underline{\textbf{32.1}} & 16.0 & \underline{\textbf{47.8}} & \underline{\textbf{36.7}} &\underline{\textbf{44.2}}\\
                    IMENet-2DGT (Ours)*  &90.0 & 81.0 &74.3 &45.6 &94.1 &43.4 &33.3 &47.6 &62.4 &61.5 &42.0 &35.9 &59.5 &46.7 & 52.0\\
					\hline
			\end{tabular}}
		\end{center}
        \caption{Semantic scene completion results on the NYU dataset with TorchSSC as our 3D backbone, and '*' are RGB-D based methods.
        }
		\label{tab:complt_nyu}
	\end{table*}
\subsection{Qualitative Analysis}
\begin{figure}[htp]
	\center
	\includegraphics[width=1\columnwidth, height=7.0cm]{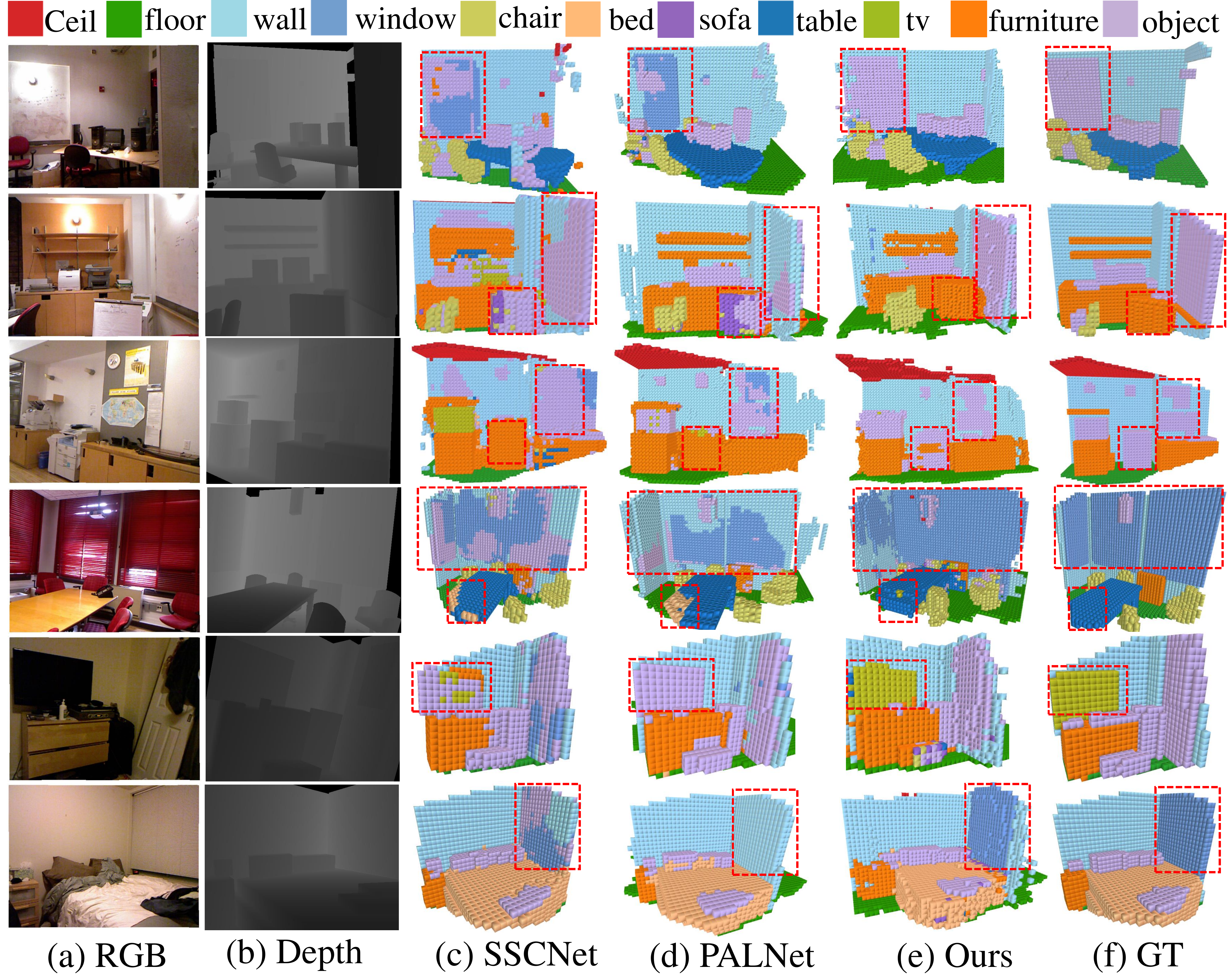}
	\caption[stereo]{\textbf{Qualitative results on the NYUCAD dataset}. From left to right: (a-b) Input RGB-D images, (c) results obtained by SSCNet \cite{song2017semantic}, (d) results generated by PALNet \cite{li2019rgbd}, (e) the proposed approach, and (f) ground truth.
}
	\label{fig-qualitive1}
\end{figure}

The visualization of some results on the NYUCAD dataset are shown in Figure \ref{fig-qualitive1} to prove the effectiveness of our IMENet qualitatively.
Firstly, our proposed method can handle diverse categories with different sizes or various shapes.
For example, whether it is a large object \emph{window} (row 4,6) or a small object \emph{tv} (row 5), our model can handle it well. 
The reason lies in that the proposed DCP may adaptively adjust the receptive field voxel-wisely.
Besides, for the categories with severe shape variability, such as \emph{table} in row 4 and  \emph{object} in row 3 right rectangle, both SSCNet, and PALNet cannot acquire satisfying shapes. However, our IMENet can model the shape boundary very well, and we attribute this to the complementary color and texture information from the 2D SS branch.
Furthermore, our method can also maintain segmentation consistency and perform better on challenging classes than SSCNet and PALNet. 
The \emph{object} in row 1 and the right rectangle of row 2, and the \emph{window} in row 4 demonstrate the superiority of our approach in the segmentation consistency.
Within the left red rectangle in row 2 and 3, both SSCNet and PALNet fail to identify the correct category (\emph{furniture} and \emph{object} misidentify each other), or they misidentify \emph{tv} in row 5.
But our method still successfully identifies these objects.
IMENet can avoid segmentation errors because it has learned the semantic priors from the 2D SS branch.
\begin{table}[htp]
\centering
\small
    \begin{center}
    \resizebox{0.35\textwidth}{!}{
        \begin{tabular}{c|ccc|c|cc}
        	\hline
        	Model & IL & DCP & DDA  & SS & SSC & SC\\
        	\hline
            A & \xmark & \xmark & \xmark
            & 46.8 & 40.3 & 72.3\\
        	B & \cmark & \xmark & \xmark
        	& 48.1 & 42.5 & 74.3\\
        	C & \cmark & \cmark &  \xmark
        	& 62.2 &42.8 & 75.5 \\
        	D & \cmark & \xmark& \cmark
        	& 47.4 & 44.8 & 77.7\\   
        	E & \cmark & \cmark & \cmark
            & \bf{65.6} &\bf{47.5} &\bf{79.1}\\
        	\hline
        \end{tabular}}
    \end{center}
    \caption{Ablation study on IL/DCP/DDA on the NYUCAD dataset.}
    \label{tab:ablation}
\end{table}
\subsection{Ablation Study}
\textbf{Different Modules in the Framework}
We first conduct ablation studies on different modules in our IMENet, as shown in Table \ref{tab:ablation}.
The baseline (model A) is set to learn SSC and SS separately without iterative learning (IL). Model A only gets
mIoU of 46.8\% on SS and 40.3\% on SSC.
Model B with IL means that we perform iterative learning several times, but there is no feature interaction between the two branches.
The SS performance of model B is slightly improved to 48.1\%, and both SC and SSC are about 2\% higher than model A.  
Moreover, Model C and D are used to verify the effectiveness of the proposed DCP and DDA modules. After the interactive features are applied via DCP, SS is greatly improved to 62.2\% by a large margin (14.1\%).
Similarly, improvements of 2.3\% and 3.4\% on SSC and SC are further obtained through DDA.
Furthermore, with both DCP and DDA added to the overall pipeline, our IMENet achieves the best results on both tasks, which are presented in bold numbers.

In addition, the 2D SS accuracy comparisons are shown in Table \ref{tab:2dss}. 
We select Seg2DNet-RGB \cite{liu2018see} as our 2D backbone of all the experiments in this paper because it released the pre-trained model. When 3D SSC features with geometric priors flow into the 2D branch via the DCP module, 2D SS predictions are significantly improved compared with RGB-D as input directly (denoted by Seg2DNet-RGBD). 
Deeplab v3+ is pretrained on ADE-20k \cite{Zhou_2017_CVPR} and finetuned on NYU with a CRF applied on the output.
Although the average IoU of our method is slightly lower than Deeplab v3+, it is much higher than other methods without any post-processing. Furthermore, IMENet can achieve the best results in some challenging categories (e.g., \emph{chair}). 
\begin{table}[htp]
		\small
		\renewcommand\tabcolsep{2.5pt}
		\begin{center}
			\resizebox{0.409\textwidth}{!}{%
				\begin{tabular}{l|ccc|c}
					\hline
                    &
					\multicolumn{3}{c|}{Selected 3 classes}
                    &11 classes \\
					Methods&	
					{wall}&
					{chair}&
					{objs.}&
					{avg. IoU} \\
					\hline
                    Deeplab v2 \cite{garbade2019two} &76.6 &58.5 &54.7 &61.5\\   %
					Deeplab v3+ \cite{garbade2019two}  &82.8  &65.8 &62.9 &\textbf{69.3}\\ %
                    \hline
                    Seg2DNet \cite{liu2018see} &54.3  &42.8  &39.8 & 46.8 \\ %
                    Seg2DNet-RGBD \cite{liu2018see} &58.8  &44.7  &42.3  &48.3   \\
                    IMENet-SSC (Ours)     &\textbf{86.0} &\textbf{77.0}  &\textbf{76.1} & \underline{65.6}\\
                    IMENet-3DGT (Ours)    &86.6 & 90.6 &98.6 &80.3 \\
					\hline
			\end{tabular}
            }
		\end{center}
        \caption{2D Semantic segmentation results on the NYU dataset. In both cases of the upper part, 
        the pre-trained models on other datasets and a CRF are used. \underline{Underlines} are higher than backbone methods.} 
		\label{tab:2dss}
	\end{table}

\begin{figure}[t]
    \center
	\includegraphics[width=0.72\columnwidth, height=3.2cm]{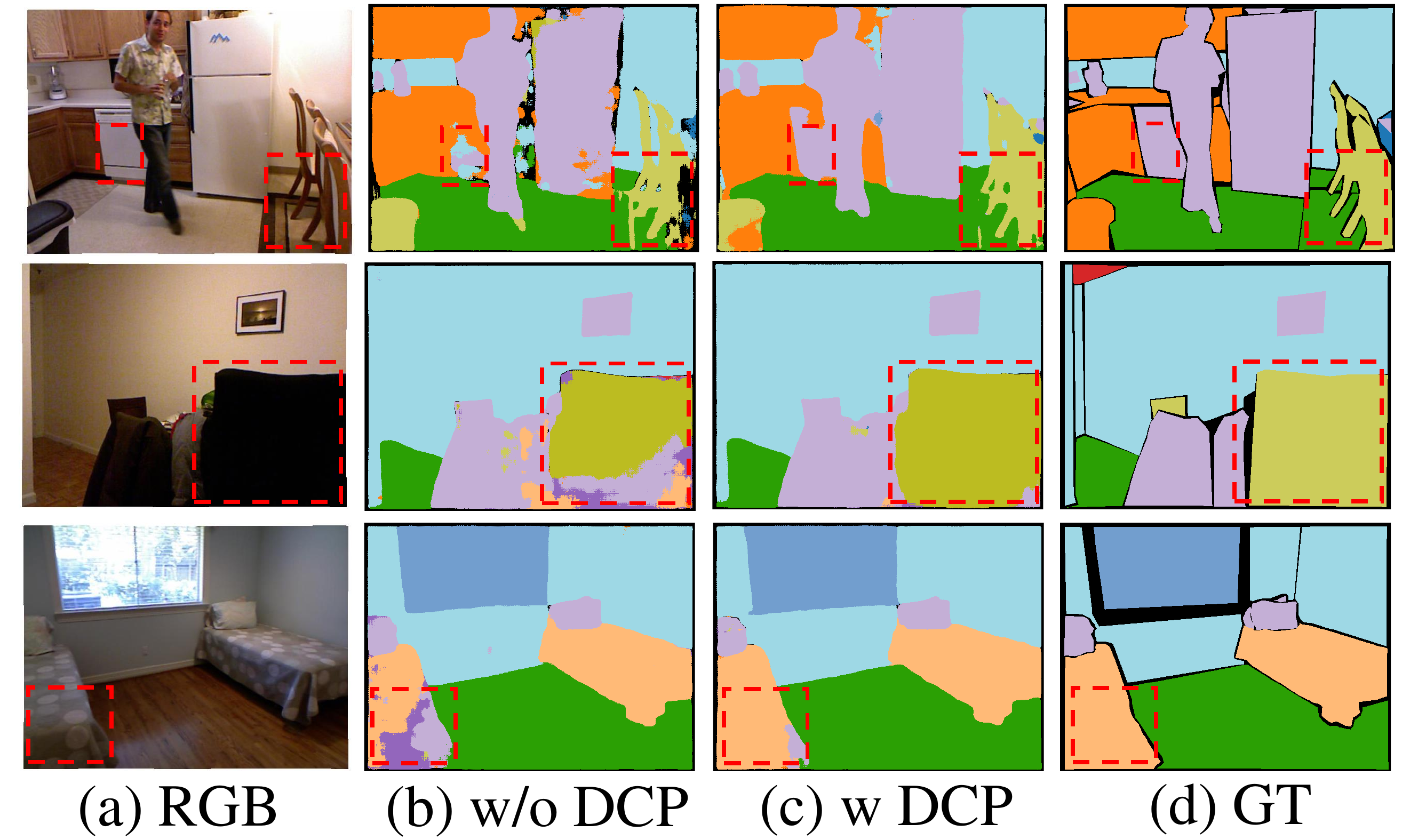}
    \vspace{-0.3cm}
	\caption[stereo]{Qualitative comparison between our approach w/o and w/ DCP on the NYU validation set.}
	\label{fig-2dss}
    \center
	\includegraphics[width=0.70\columnwidth, height=3.0cm]{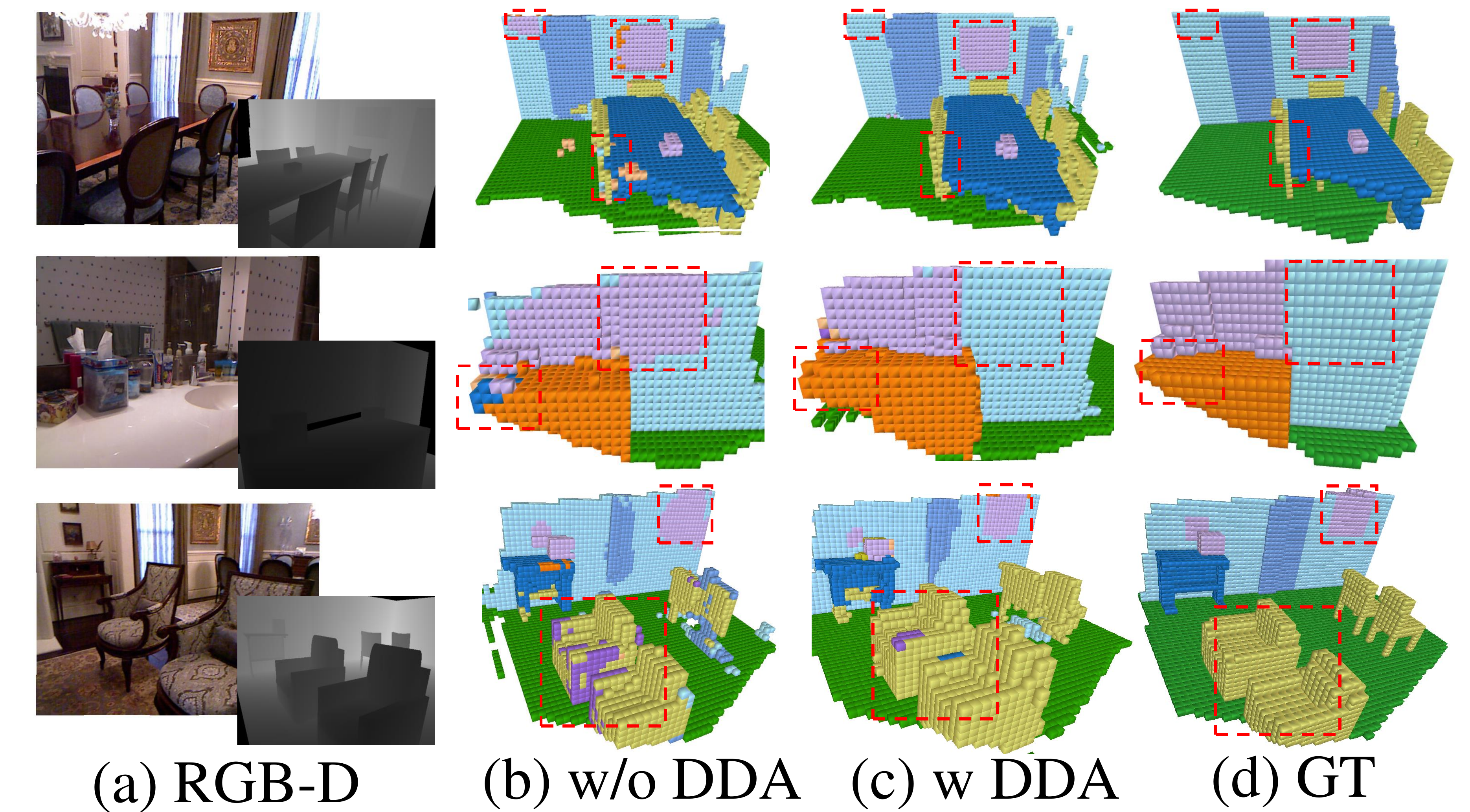}
    \vspace{-0.3cm}
	\caption[stereo]{Qualitative comparison between our approach w/o and w/ DDA on the NYUCAD validation set.}
	\label{fig-3dssc}
\vspace{-0.4cm}
\end{figure}

\textbf{Impacts of DCP and DDA}
To further validate the effectiveness of DCP and DDA, the comparison results are shown in Figure \ref{fig-2dss} and Figure \ref{fig-3dssc}, respectively.
We can observe that DCP can successfully identify different categories, even if their colors/textures are very similar, as shown in red squares.
Take the first row in Figure \ref{fig-2dss} as an example, because the colors/textures of \emph{chair} and \emph{floor} are similar, the IMENet (w/o DCP) will easily mislabel \emph{floor} as \emph{chair}, while IMENet (w DCP) can rectify this error.
We attribute this to the DCP's ability to effectively integrate shape/depth geometries introduced from the 3D branch.
Similarly, DDA can scrutinize more discriminative contexts even though they have similar depths, and
we owe it to the advantage that DDA can adequately fuse the colors/textures obtained from the 2D branch.  

\begin{figure}[htp]
	\center
	\includegraphics[width=0.45\columnwidth, height=3.0cm]{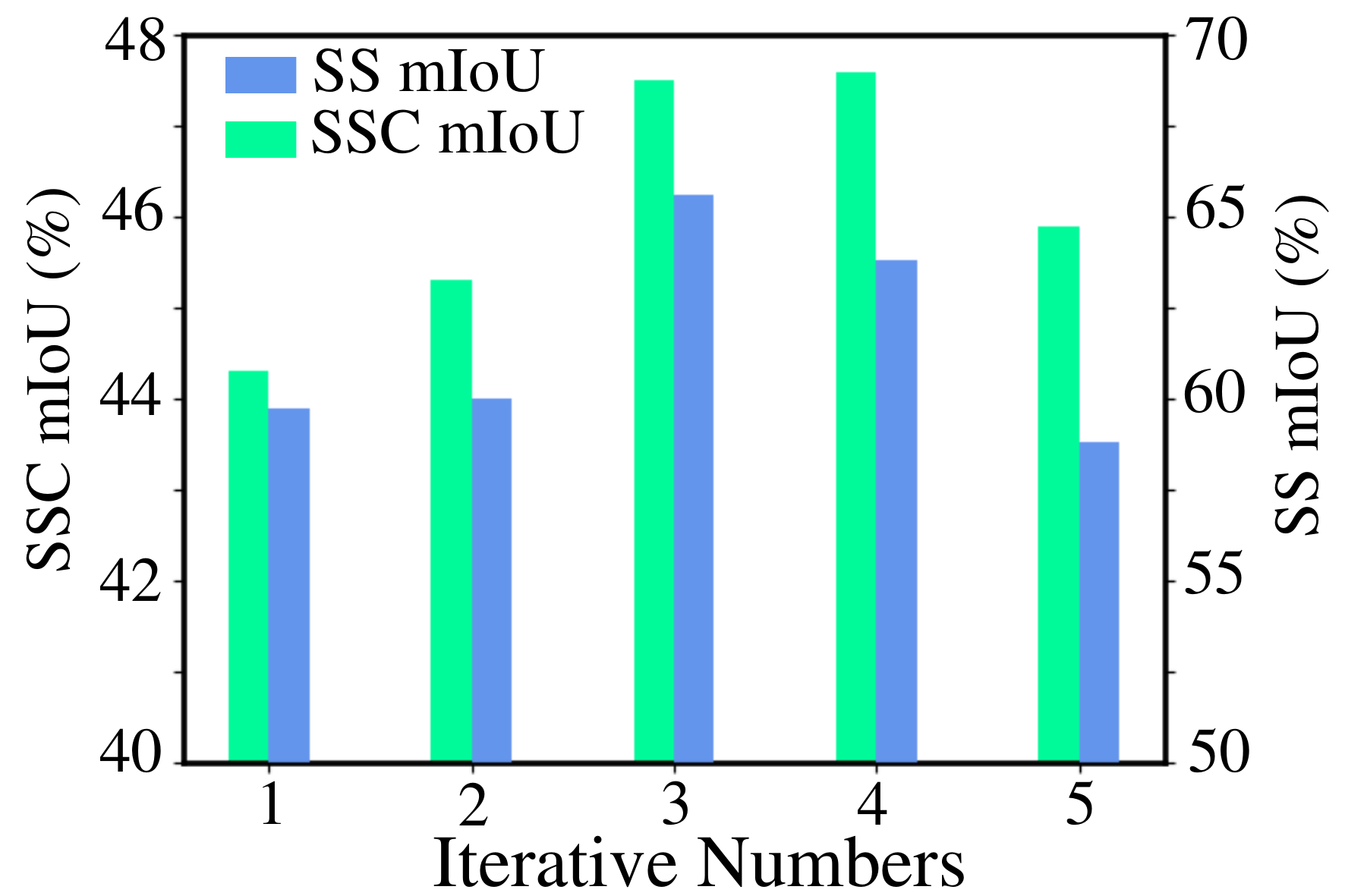}
    \vspace{-0.2cm}
	\caption[stereo]{Accuracies of SSC and SS with different iterative times.}
	\label{fig-iternums}
    \vspace{-0.2cm}
\end{figure}

\textbf{Analysis of Iterative Numbers}
Experiments are conducted to analyze the impact of different iterative numbers, and the accuracies of both SSC and SS tasks are shown in Figure \ref{fig-iternums}.
It can be observed that the results first increase with more iterations, then they get stable around 3 and 4. Further increasing the number of iterations degrades the performance. This phenomenon is consistent with the observation of \cite{zhang2019joint}.
The network convergence may explain this phenomenon. In general, using more iterations can improve the performance to some extent.
But more iterations also require more training for the network to converge.\vspace{-0.2cm}
\vspace{-0.15cm}
\section{Conclusion}
This paper proposes an iterative mutual enhancement network named IMENet to further exploit bi-directional fusions between semantic completion and segmentation tasks via two task-shared refined modules, i.e., Deformable Context Pyramid and Deformable Depth Attention.
By leveraging the late fusion method, our IMENet can fully explore specific high-level features of multi-modality input. 
Finally, elaborate experiments on both NYU and NYUCAD datasets show that our proposed IMENet achieves state-of-the-art results on both semantic scene completion and semantic segmentation tasks.

\section*{Acknowledgments}
This work was supported in part by Shenzhen Natural Science Foundation under Grant JCYJ20190813170601651, and in part by funding from Shenzhen Institute of Artificial Intelligence and Robotics for Society.
\clearpage
\newpage
\small
\bibliographystyle{named}
\bibliography{ijcai21}

\end{document}